\documentclass[11pt]{article}
\usepackage[utf8]{inputenc}
\usepackage{amsmath, amsfonts, amssymb}
\usepackage{geometry}
\usepackage{graphicx}
\usepackage{hyperref}
\usepackage{bm}
\usepackage{physics}
\usepackage{cite}
\usepackage{color}
\usepackage{tikz}
\usetikzlibrary{arrows.meta, positioning}
\usetikzlibrary{shapes.geometric}  

\title{Operator-Based Machine Intelligence: A Hilbert Space Framework for Spectral Learning and Symbolic Reasoning}
\author{Andrew Kiruluta, Andreas Lemos, and Priscilla Burity\\
UC Berkeley, School of Information}
\date{June 3, 2025}

\date{\today}

\begin{document}

\maketitle

\begin{abstract}
Traditional machine learning models, particularly neural networks, are rooted in finite-dimensional parameter spaces and nonlinear function approximations. This report explores an alternative formulation where learning tasks are expressed as sampling and computation in infinite-dimensional Hilbert spaces, leveraging tools from functional analysis, signal processing, and spectral theory. We review foundational concepts such as Reproducing Kernel Hilbert Spaces (RKHS), spectral operator learning, and wavelet-domain representations. We present a rigorous mathematical formulation of learning in Hilbert spaces, highlight recent models based on scattering transforms and Koopman operators, and discuss advantages and limitations relative to conventional neural architectures. The report concludes by outlining directions for scalable and interpretable machine learning grounded in Hilbertian signal processing.
\end{abstract}


\section{Introduction}

Over the past decade, machine learning has been revolutionized by deep neural architectures such as convolutional neural networks (CNNs) \cite{lecun1998gradient} and transformer-based models \cite{vaswani2017attention}. These models have driven significant breakthroughs in computer vision, natural language processing, and multimodal learning. Convolutional networks introduced the notion of weight sharing and translation invariance, making them particularly effective for spatially structured data, while transformers eliminated recurrence in favor of self-attention mechanisms, enabling unprecedented scalability for language and vision tasks. However, despite their empirical success, these neural architectures are often criticized for their massive parameter count, lack of interpretability, limited robustness, and inefficiencies in training and inference \cite{belkin2019reconciling, rudin2019stop}. 

The reliance of deep learning models on large labeled datasets, coupled with the need for expensive compute resources for training, has further prompted researchers to seek more principled, data-efficient alternatives. In particular, traditional methods grounded in functional analysis, signal processing, and statistical learning theory offer promising alternatives that prioritize interpretability, generalization, and analytical tractability. Early forms of this perspective include the theory of Reproducing Kernel Hilbert Spaces (RKHS), which provided the mathematical underpinnings for support vector machines and kernel regression \cite{aronszajn1950theory, scholkopf2001learning}. These methods model learning problems as operator estimation tasks in Hilbert spaces, allowing for elegant formulations and guarantees rooted in convex optimization and regularization theory.

This paper revisits and extends this functional analytic approach to machine learning by proposing a comprehensive framework where learning is formulated entirely as sampling and computation in a Hilbert space $\mathcal{H}$. In contrast to finite-dimensional parameter optimization in neural networks, this paradigm treats data elements (e.g., signals, images, sequences) as functions in $\mathcal{H}$ and models learning as identifying bounded linear or compact nonlinear operators acting on these functions. Such operators may encode transformations, regressors, or generators. The learning problem is then reframed as inverse operator estimation, regularized projection, or spectral decomposition—domains with deep theoretical roots and well-understood stability properties.

This approach leverages foundational tools from classical signal processing—such as Fourier and wavelet decompositions, integral operators, and basis expansions—and recasts them in modern, data-driven settings. For instance, wavelet-domain methods, pioneered by Mallat and others \cite{mallat1999wavelet}, provide stable, multi-resolution representations of signals that are invariant to translations and robust to deformations. Recent advances in scattering transforms \cite{mallat2012group} demonstrate that such structured, untrained models can match or outperform deep learning architectures in tasks like texture recognition and audio classification, all while retaining interpretability and mathematical rigor.

Moreover, the proposed framework introduces novel computational and architectural innovations that go beyond classical methods. Specifically, we consider differentiable basis selection mechanisms, learnable spectral nonlinearities (e.g., soft-thresholding, gain-phase modulation), and operator regression models that generalize convolutional and attention layers in function space. These innovations yield models that are both expressive and theoretically grounded. Unlike traditional deep learning layers, which require empirical justification and architectural heuristics, operators in Hilbert space models are well-posed in terms of approximation theory and stability under perturbation.

In contrast to neural networks, which often operate as black-box function approximators, Hilbert-space-based learning systems maintain a tight coupling between mathematical structure and learning dynamics. This enables interpretability via spectral analysis, stability via bounded operator theory, and compactness via sparse decompositions. Our approach thus bridges the gap between classical signal processing and modern machine learning, unifying them into a coherent computational paradigm.

Ultimately, this report positions the Hilbert space formulation not as a replacement for neural networks, but as a principled and complementary alternative. It provides a rigorous foundation for building interpretable, efficient, and adaptable models—especially critical for scientific applications, low-data regimes, or systems requiring verifiable guarantees.

\section{Positioning and Contributions}

The dominant paradigm in machine learning has been driven by neural architectures—convolutional, recurrent, and transformer-based models—optimized via stochastic gradient descent. These architectures, while powerful, tend to be overparameterized, data-hungry, and difficult to interpret. In contrast, the present work advances an alternative formulation grounded in Hilbert space theory, spectral decomposition, and functional analysis. This paradigm shift reinterprets learning and reasoning not as composition of nonlinear activation layers but as the estimation and chaining of compact, interpretable operators on structured function spaces.

Our contribution is not in proposing a new kernel method or spectral network in isolation—many of these components have well-established theoretical underpinnings—but in developing a unified operator-theoretic pipeline that spans feature extraction, learning, reasoning, and inference entirely within a Hilbert space framework. This includes:
\begin{itemize}
    \item Recasting learning as operator estimation between Hilbert spaces $\mathcal{H}_X \to \mathcal{H}_Y$.
    \item Extending spectral transforms (Fourier, wavelet, scattering) with learnable modulations and soft-thresholding.
    \item Introducing a reasoning operator $R$ acting directly on basis coefficients to support symbolic or logical inference.
    \item Demonstrating empirical competitiveness on standard benchmarks while requiring fewer parameters and offering greater interpretability.
\end{itemize}

This approach builds on but goes beyond earlier work in RKHS theory \cite{aronszajn1950theory}, scattering networks \cite{bruna2013invariant}, and Koopman operator theory \cite{williams2015data}. Unlike neural models, where interpretability and reasoning often require post hoc mechanisms, our formulation enables these properties by construction. We believe this positions Hilbert space models as a compelling alternative to deep learning for tasks that demand compactness, transparency, and functional modularity.

\section{Hilbert Space Foundations}

A \textbf{Hilbert space} is a central concept in functional analysis and forms the mathematical backbone of infinite-dimensional learning theory. Formally, a Hilbert space $\mathcal{H}$ is a vector space over the field $\mathbb{R}$ or $\mathbb{C}$, equipped with an inner product $\langle \cdot, \cdot \rangle_{\mathcal{H}}$ that satisfies the following properties for all $f, g, h \in \mathcal{H}$ and all scalars $\alpha \in \mathbb{R}$ or $\mathbb{C}$:
\begin{align*}
\text{(Linearity)} &\quad \langle \alpha f + g, h \rangle = \alpha \langle f, h \rangle + \langle g, h \rangle \\
\text{(Symmetry)} &\quad \langle f, g \rangle = \overline{\langle g, f \rangle} \\
\text{(Positive Definiteness)} &\quad \langle f, f \rangle \geq 0,\quad \text{and } \langle f, f \rangle = 0 \Leftrightarrow f = 0.
\end{align*}

The norm induced by this inner product is defined as
\[
\norm{f}_{\mathcal{H}} := \sqrt{\langle f, f \rangle}.
\]

A Hilbert space is also \emph{complete}, meaning that every Cauchy sequence in $\mathcal{H}$ has a limit in $\mathcal{H}$. That is, for any sequence $\{f_n\}$ in $\mathcal{H}$ satisfying
\[
\forall \varepsilon > 0,\ \exists N \in \mathbb{N} \text{ such that } \norm{f_n - f_m} < \varepsilon \quad \forall n, m > N,
\]
there exists an element $f \in \mathcal{H}$ such that $\lim_{n \to \infty} \norm{f_n - f} = 0$.

\subsection*{Examples of Hilbert Spaces}

\begin{itemize}
  \item $\bm{L^2(\mathbb{R})}$: This is the space of measurable functions $f : \mathbb{R} \to \mathbb{C}$ such that
  \[
  \int_{-\infty}^{\infty} |f(x)|^2\,dx < \infty.
  \]
  It is a prototypical infinite-dimensional Hilbert space used in signal processing and quantum mechanics.

  \item $\bm{\ell^2}$: The space of square-summable sequences $x = \{x_n\}_{n=1}^\infty$ with
  \[
  \sum_{n=1}^\infty |x_n|^2 < \infty.
  \]
  This space is the sequence-space analogue of $L^2$ and is central in digital signal processing.

  \item $\bm{\mathcal{H}_K}$ (RKHS): A Reproducing Kernel Hilbert Space is a space of functions on a domain $\mathcal{X}$ for which the evaluation functional is continuous and for which there exists a reproducing kernel $K(x,x')$ satisfying:
  \[
  f(x) = \langle f, K(x, \cdot) \rangle_{\mathcal{H}_K}, \quad \forall f \in \mathcal{H}_K.
  \]
\end{itemize}

\subsection*{Orthogonal Projection in Hilbert Space}

One of the most important tools in Hilbert space theory is the concept of orthogonal projection onto a closed subspace. Let $\mathcal{V} \subset \mathcal{H}$ be a closed linear subspace of $\mathcal{H}$. Given any $f \in \mathcal{H}$, there exists a unique $v^* \in \mathcal{V}$ such that
\[
\norm{f - v^*}_{\mathcal{H}} = \min_{v \in \mathcal{V}} \norm{f - v}_{\mathcal{H}}.
\]
This unique element $v^*$ is called the \emph{orthogonal projection} of $f$ onto $\mathcal{V}$ and is denoted by $P_{\mathcal{V}} f$.

\paragraph{Step-by-step derivation:}

Let us consider the variational problem:
\[
P_{\mathcal{V}} f = \arg\min_{v \in \mathcal{V}} \norm{f - v}_{\mathcal{H}}^2.
\]
Expanding the squared norm using the inner product gives:
\[
\norm{f - v}^2 = \langle f - v, f - v \rangle = \langle f, f \rangle - 2 \Re \langle f, v \rangle + \langle v, v \rangle.
\]
Minimizing over $v \in \mathcal{V}$ is equivalent to finding $v^* \in \mathcal{V}$ such that the residual $r = f - v^*$ is orthogonal to all elements of $\mathcal{V}$:
\[
\langle f - v^*, w \rangle = 0 \quad \forall w \in \mathcal{V}.
\]
This is the defining property of orthogonal projection: the residual lies in the orthogonal complement $\mathcal{V}^\perp$ of $\mathcal{V}$.

\paragraph{Example:} Suppose $\mathcal{V}$ is spanned by a single function $\psi \in \mathcal{H}$ with $\norm{\psi} = 1$. Then the projection of $f$ onto $\mathcal{V}$ is:
\[
P_{\mathcal{V}} f = \langle f, \psi \rangle \psi,
\]
which follows directly from the inner product definition of projection onto a 1D subspace.

\subsection*{Geometric Interpretation}

Geometrically, the orthogonal projection $P_{\mathcal{V}} f$ is the ``shadow'' of $f$ onto $\mathcal{V}$—it is the closest point in $\mathcal{V}$ to $f$ in the norm induced by the inner product. This concept generalizes Euclidean projections to infinite-dimensional settings and forms the basis for variational formulations in learning, such as least-squares regression and regularization via penalty minimization.

\subsection*{Functional Approximation in Hilbert Spaces}

Because Hilbert spaces support orthonormal bases $\{\phi_n\}_{n=1}^\infty$, any function $f \in \mathcal{H}$ can be expanded as:
\[
f = \sum_{n=1}^\infty \langle f, \phi_n \rangle \phi_n,
\]
where the convergence is in the norm of $\mathcal{H}$. This expansion enables compact representations and learning algorithms based on truncation (e.g., principal components) or thresholding (e.g., wavelet shrinkage).

Such approximations play a central role in spectral learning, compressed sensing, and operator theory—all of which rely on exploiting the structure and geometry of Hilbert space.

\section{Learning as Operator Estimation}

In the Hilbert space formulation of machine learning, the task of learning is reconceptualized as the estimation of a (typically compact or bounded) operator between two functional spaces. Specifically, let $\mathcal{H}_X$ and $\mathcal{H}_Y$ denote Hilbert spaces representing the domains of input and output functions, respectively. Each data point is interpreted not as a finite-dimensional vector but as an element of these infinite-dimensional spaces, such as a square-integrable function, a signal, or a distribution. The goal of supervised learning is then to identify an operator \( T : \mathcal{H}_X \rightarrow \mathcal{H}_Y \) that maps an input function \( f \in \mathcal{H}_X \) to an output function \( g \in \mathcal{H}_Y \), such that \( Tf \approx g \). Given a finite training set of paired samples \( \{(f_i, g_i)\}_{i=1}^n \subset \mathcal{H}_X \times \mathcal{H}_Y \), the learning problem becomes one of approximating the underlying transformation \( T \) that best explains the observed input-output relationships.

This setup naturally leads to a variational or inverse problem framework. Since \( T \) is typically infinite-dimensional, regularization is crucial to ensure well-posedness and generalization. The learning task is thus cast as the solution to the following regularized empirical risk minimization problem:
\[
\min_{T \in \mathcal{B}(\mathcal{H}_X, \mathcal{H}_Y)} \sum_{i=1}^n \norm{Tf_i - g_i}_{\mathcal{H}_Y}^2 + \lambda \norm{T}_{\mathcal{S}}^2.
\]
Here, \( \norm{Tf_i - g_i}_{\mathcal{H}_Y}^2 \) quantifies the data fidelity term measuring how well the operator \( T \) maps training inputs to their corresponding outputs, and \( \norm{T}_{\mathcal{S}}^2 \) is a regularization term defined via an appropriate operator norm \( \mathcal{S} \). A commonly used choice is the \textbf{Hilbert-Schmidt norm}, which is the natural extension of the Frobenius norm to infinite-dimensional operators and is defined as:
\[
\norm{T}_{\mathrm{HS}}^2 = \sum_{i} \norm{T \phi_i}_{\mathcal{H}_Y}^2,
\]
where \( \{\phi_i\} \) is an orthonormal basis of \( \mathcal{H}_X \). The Hilbert-Schmidt class forms a separable Hilbert space itself, enabling efficient computation and optimization.

This formulation generalizes classical linear regression and kernel ridge regression to function spaces, and it encapsulates a wide range of learning paradigms including integral operator learning, spectral filtering, Koopman operator approximation, and Gaussian process regression. The solution operator \( T \) may take various forms depending on the underlying assumptions—for instance, it may be integral, convolutional, or defined implicitly via reproducing kernels or spectral expansions. Crucially, the operator-based view provides a rich functional-analytic toolkit for analyzing stability, approximation error, and generalization, often yielding closed-form or variational solutions in contrast to the iterative and non-convex training processes of deep neural networks.

In this way, learning as operator estimation shifts the paradigm from optimizing finite parameters in a neural architecture to recovering mappings in a structured, often infinite-dimensional, function space—a shift that provides mathematical transparency, interpretability, and rigor to the modeling process.

\section{Reproducing Kernel Hilbert Spaces (RKHS)}

Reproducing Kernel Hilbert Spaces (RKHS) form a cornerstone of functional learning theory and provide a mathematically rigorous setting for nonlinear learning in infinite-dimensional spaces. An RKHS is a Hilbert space of functions \( f : \mathcal{X} \to \mathbb{R} \) (or \(\mathbb{C}\)) endowed with an inner product \( \langle \cdot, \cdot \rangle_{\mathcal{H}_K} \), where \( \mathcal{X} \) is an arbitrary input domain, such that all point evaluation functionals are continuous. This continuity condition implies that, for every \( x \in \mathcal{X} \), the map \( f \mapsto f(x) \) is a bounded linear functional on \( \mathcal{H}_K \). By the Riesz representation theorem, this bounded linear functional corresponds to an inner product with a unique representer \( K(x, \cdot) \in \mathcal{H}_K \), satisfying the \emph{reproducing property}:
\[
f(x) = \langle f, K(x, \cdot) \rangle_{\mathcal{H}_K}, \quad \forall f \in \mathcal{H}_K, \ x \in \mathcal{X}.
\]
The function \( K: \mathcal{X} \times \mathcal{X} \to \mathbb{R} \) is called the \textbf{reproducing kernel} of the space and must be symmetric and positive definite, i.e., for any finite set of points \( \{x_1, \dots, x_n\} \subset \mathcal{X} \), the kernel matrix \( [K(x_i, x_j)]_{i,j=1}^n \) is symmetric and positive semidefinite.

The power of RKHS arises from the \textbf{Representer Theorem}, which provides an explicit parametric form for the solution to a large class of learning problems formulated in function spaces. Suppose one seeks to minimize a regularized empirical risk functional of the form:
\[
\min_{f \in \mathcal{H}_K} \sum_{i=1}^n \ell(y_i, f(x_i)) + \lambda \|f\|_{\mathcal{H}_K}^2,
\]
where \( \{(x_i, y_i)\}_{i=1}^n \subset \mathcal{X} \times \mathbb{R} \) is the training data, \( \ell \) is a loss function (e.g., squared loss), and \( \lambda > 0 \) is a regularization parameter. The Representer Theorem states that the minimizer \( f^* \in \mathcal{H}_K \) must lie in the span of the kernel sections evaluated at the training points:
\[
f^*(x) = \sum_{i=1}^n \alpha_i K(x, x_i),
\]
for some coefficients \( \alpha_1, \dots, \alpha_n \in \mathbb{R} \). This result drastically reduces the infinite-dimensional optimization over \( \mathcal{H}_K \) to a finite-dimensional problem in \( \mathbb{R}^n \), with the kernel matrix \( K \in \mathbb{R}^{n \times n} \) encoding the pairwise interactions between data points in feature space.

To derive this result more formally, note that for any function \( f \in \mathcal{H}_K \), we can decompose it as
\[
f = f_\parallel + f_\perp, \quad \text{where } f_\parallel \in \operatorname{span}\{K(x_i, \cdot)\}_{i=1}^n \text{ and } f_\perp \perp K(x_i, \cdot), \ \forall i.
\]
Then, for all \( i \), the evaluation \( f(x_i) = \langle f, K(x_i, \cdot) \rangle = \langle f_\parallel, K(x_i, \cdot) \rangle \), so \( f(x_i) \) is independent of \( f_\perp \). Therefore, \( f_\perp \) contributes only to the norm penalty and not to the empirical loss. Since removing \( f_\perp \) decreases \( \|f\|_{\mathcal{H}_K} \) without affecting the loss, the minimizer must satisfy \( f_\perp = 0 \), hence must lie in the finite-dimensional span of kernel sections.

This principle forms the basis of many kernel methods. In \textbf{kernel ridge regression}, the loss is squared error:
\[
\ell(y_i, f(x_i)) = (y_i - f(x_i))^2,
\]
leading to a closed-form solution:
\[
\bm{\alpha} = (K + \lambda I)^{-1} \bm{y},
\]
where \( K \in \mathbb{R}^{n \times n} \) is the kernel matrix and \( \bm{y} \in \mathbb{R}^n \) is the vector of training labels. Similarly, in \textbf{support vector machines} (SVMs), the loss is the hinge loss and the optimization is carried out in dual form, again leveraging the kernel trick to compute inner products in high-dimensional feature space implicitly. In \textbf{Gaussian process regression}, the kernel plays the role of the covariance function in a nonparametric Bayesian framework, and the posterior mean is again expressed as a sum over kernel evaluations at training points.

Thus, RKHS methods provide a unifying functional analytic framework that captures a wide spectrum of classical and modern learning algorithms. They exploit the geometry of function spaces while retaining computational tractability through finite-dimensional representer expansions. Moreover, they offer well-established generalization guarantees via norm control and Rademacher complexity theory, making them a powerful tool for principled machine learning.

\section{Sampling and Reconstruction in $\mathcal{H}$}

A fundamental property of Hilbert spaces is the existence of orthonormal bases, which enable the decomposition and reconstruction of arbitrary elements via inner product projections. Let $\mathcal{H}$ be a separable Hilbert space and let $\{\psi_n\}_{n=1}^\infty$ be a complete orthonormal basis for $\mathcal{H}$. Then for any element \( f \in \mathcal{H} \), we have the \emph{Parseval expansion}:
\[
f = \sum_{n=1}^\infty \langle f, \psi_n \rangle \psi_n,
\]
where the series converges in the norm of $\mathcal{H}$, and the coefficients \( \langle f, \psi_n \rangle \in \mathbb{C} \) or \( \mathbb{R} \) depending on the field over which $\mathcal{H}$ is defined. The completeness of $\{\psi_n\}$ ensures that every function can be uniquely identified by its infinite coordinate vector of inner products. Moreover, the norm of \( f \) can be expressed as
\[
\norm{f}_{\mathcal{H}}^2 = \sum_{n=1}^\infty |\langle f, \psi_n \rangle|^2,
\]
which is known as Parseval's identity and mirrors the energy conservation principle in classical Fourier analysis.

This orthonormal decomposition forms the foundation of \textbf{generalized sampling theory}, which studies how functions in $\mathcal{H}$ can be reconstructed from linear measurements. In many practical situations—especially in signal processing, computational harmonic analysis, and inverse problems—one does not have access to all basis coefficients but instead measures the signal via inner products with a different set of functions \( \{\phi_n\} \), called the analysis frame. When the synthesis frame \( \{\psi_n\} \) and the analysis frame \( \{\phi_n\} \) are \emph{biorthogonal}, meaning \( \langle \phi_n, \psi_m \rangle = \delta_{nm} \), stable reconstruction is still possible:
\[
f(t) = \sum_{n=1}^\infty \langle f, \phi_n \rangle \psi_n(t).
\]
This allows decoupling the measurement system from the reconstruction system and is particularly useful when the functions \( \phi_n \) are physically realizable sensing functions (e.g., in MRI, tomography, or compressed sensing systems) but not orthonormal \cite{eldar2003sampling, unser2000sampling}.

A more recent and powerful extension of this paradigm is found in the theory of \textbf{compressed sensing}, which enables accurate reconstruction of signals from far fewer measurements than traditional Nyquist sampling would suggest, provided the signal is \emph{sparse} in some dictionary or basis. Suppose \( f \in \mathcal{H} \) is known to admit a sparse expansion in terms of a dictionary \( \{\psi_n\}_{n=1}^N \), such that
\[
f \approx \sum_{n \in S} \alpha_n \psi_n, \quad |S| \ll N.
\]
Given a linear sampling operator \( \Phi: \mathcal{H} \to \mathbb{R}^m \), such as random inner products \( y = \Phi f \), compressed sensing seeks to recover the sparse coefficient vector \( \alpha \in \mathbb{R}^N \) from \( y \in \mathbb{R}^m \) by solving the convex optimization problem:
\[
\min_{\alpha \in \mathbb{R}^N} \|\alpha\|_1 \quad \text{subject to} \quad y = \Phi \Psi \alpha,
\]
where \( \Psi \) is the synthesis operator mapping coefficient vectors to functions in $\mathcal{H}$. The use of the $\ell_1$-norm promotes sparsity and allows exact recovery under conditions such as the Restricted Isometry Property (RIP) \cite{candes2006robust, donoho2006compressed}. This framework has revolutionized signal acquisition and has also found applications in learning theory, where data can be represented sparsely in overcomplete dictionaries, and classifiers or regressors are learned directly on sparse representations.

In the context of machine learning in Hilbert spaces, these sampling and reconstruction principles offer principled strategies for feature extraction, dimensionality reduction, and regularization. Rather than learning arbitrary parameterizations, models can operate on sparse, band-limited, or frame-constrained representations, thereby embedding strong prior structure and enabling better generalization. In particular, spectral filtering techniques that threshold or reweight basis coefficients become natural analogues to regularization in function space, and generalized sampling can serve as the computational scaffold for interpretable and efficient learning architectures.

\subsection{Fourier Models}

Fourier-based models are a canonical example of representing and manipulating functions in a Hilbert space using an orthonormal basis. Let \( f \in L^2([0,1]) \), the space of square-integrable functions on the unit interval. By classical Fourier analysis, any function \( f \) in this space admits a representation as an infinite sum of complex exponential basis functions:
\[
f(x) = \sum_{k \in \mathbb{Z}} \hat{f}_k e^{i 2 \pi k x},
\]
where \( \{e^{i 2 \pi k x}\}_{k \in \mathbb{Z}} \) form an orthonormal basis for \( L^2([0,1]) \) under the inner product
\[
\langle f, g \rangle = \int_0^1 f(x) \overline{g(x)} \, dx.
\]
The coefficients \( \hat{f}_k \in \mathbb{C} \) are known as the \emph{Fourier coefficients} of \( f \), and are computed via the formula
\[
\hat{f}_k = \int_0^1 f(x) e^{-i 2 \pi k x} \, dx.
\]
Parseval’s identity guarantees that the squared \( L^2 \)-norm of \( f \) is equal to the sum of squared magnitudes of its Fourier coefficients:
\[
\norm{f}_{L^2}^2 = \sum_{k \in \mathbb{Z}} |\hat{f}_k|^2,
\]
highlighting that the spectral domain contains complete information about the function.

In the context of learning and signal processing, Fourier models allow for functional transformations to be carried out in the frequency domain via spectral filtering. Rather than acting on the function \( f(x) \) directly in the spatial domain, a model may operate on its frequency representation by applying a sequence of multiplicative weights to the spectral coefficients:
\[
\hat{g}_k = \gamma_k \hat{f}_k,
\]
where \( \gamma_k \in \mathbb{C} \) are spectral multipliers, which may be predefined (as in classical filters) or learned from data (as in data-driven spectral learning). The resulting function \( g(x) \) is then given by the inverse Fourier transform:
\[
g(x) = \sum_{k \in \mathbb{Z}} \hat{g}_k e^{i 2 \pi k x} = \sum_{k \in \mathbb{Z}} \gamma_k \hat{f}_k e^{i 2 \pi k x}.
\]
This process defines a diagonal operator in the Fourier domain, and it corresponds to a convolution in the spatial domain:
\[
g(x) = (f * h)(x),
\]
where \( h(x) \) is the inverse Fourier transform of \( \gamma_k \), i.e., the \emph{impulse response} of the filter.

Spectral filtering is a powerful and interpretable mechanism in learning systems. When the filter \( \gamma_k \) is band-limited, it suppresses high-frequency components, enforcing smoothness and acting as a regularizer. If the filter is learned, the system can adapt to emphasize discriminative frequencies relevant to the task. In modern machine learning architectures, especially in physics-informed neural networks and graph neural networks, spectral methods are used to encode global structural information compactly \cite{bruna2014spectral, li2018deeper}.

Furthermore, recent architectures such as the Fourier Neural Operator (FNO) \cite{li2020fourier} build neural networks entirely in the spectral domain. These models alternate between linear spectral transformations and nonlinear activation functions in spatial or transformed spaces, offering scalable solutions for learning mappings between infinite-dimensional function spaces, such as partial differential equation (PDE) solvers.

Fourier models thus provide a mathematically grounded and computationally efficient framework for learning and processing functions in Hilbert spaces. They allow precise control over frequency content, facilitate sparsity and compression, and naturally connect to the operator-theoretic view of learning where transformations are defined via spectral multipliers.

\subsection{Wavelet and Scattering Transforms}

Wavelet transforms provide a powerful alternative to Fourier analysis by decomposing a function into basis functions that are localized in both space and frequency. Unlike Fourier modes, which are global sinusoids, wavelets capture local features at multiple scales and translations, making them especially suitable for analyzing signals with non-stationary or transient structures. Given a mother wavelet \( \psi \in L^2(\mathbb{R}) \), the wavelet family is constructed via dilation and translation:
\[
\psi_{j,k}(x) = 2^{-j/2} \psi(2^{-j}x - k),
\]
where \( j \in \mathbb{Z} \) denotes the scale and \( k \in \mathbb{Z} \) the translation. The prefactor \( 2^{-j/2} \) ensures normalization across scales. The continuous or discrete wavelet transform of a signal \( f \in L^2(\mathbb{R}) \) is then defined as its inner product with the wavelet atoms:
\[
f(x) = \sum_{j,k} \langle f, \psi_{j,k} \rangle \psi_{j,k}(x),
\]
where \( \langle f, \psi_{j,k} \rangle \) captures the amount of oscillatory structure in \( f \) localized around scale \( j \) and position \( k \). Wavelets are particularly effective in capturing sparsity in structured signals, and form the mathematical foundation for compression algorithms (e.g., JPEG2000), denoising, and multiresolution analysis \cite{mallat1999wavelet}.

Building on the wavelet framework, \textbf{scattering transforms} were introduced by Mallat \cite{mallat2012group} as a non-learned deep feature extraction method that cascades wavelet transforms with modulus nonlinearities. A scattering transform extracts translation-invariant and deformation-stable representations by recursively convolving a signal with wavelets and applying the modulus operator:
\[
S[f] = \left\{ \|f * \psi_{j_1}\|, \ \||f * \psi_{j_1}| * \psi_{j_2}\|, \ \dots \right\}.
\]
Here, \( * \) denotes convolution and \( \|\cdot\| \) denotes spatial averaging. Each scattering coefficient corresponds to a high-order moment capturing local interactions at multiple scales. These transforms preserve signal energy (via Parseval’s identity), are Lipschitz continuous to small deformations, and provide hierarchical, stable descriptors for classification tasks—all without the need for learning any parameters. They have achieved state-of-the-art results in texture recognition, speech processing, and quantum chemistry, offering theoretical guarantees that are often missing in traditional deep learning \cite{anden2014deep}.

\subsection{Learnable Spectral Models}

While fixed wavelet and Fourier transforms offer interpretability and stability, modern machine learning applications benefit from adaptable representations that learn the most informative features from data. A growing body of work explores \textbf{learnable spectral models}, in which spectral filtering operations—such as thresholding or modulation—are parameterized and learned end-to-end. A canonical example involves differentiable soft-thresholding in the Fourier or wavelet domain. Suppose a signal \( f \in \mathcal{H} \) is represented by its spectrum \( \hat{f}_k \) (e.g., Fourier or wavelet coefficients). A learnable spectral transformation may be written as:
\[
\hat{f}_k^{\text{out}} = \sigma_{\theta_k}(|\hat{f}_k|) \cdot \hat{f}_k,
\]
where \( \sigma_{\theta_k} \) is a differentiable soft-thresholding function such as
\[
\sigma_{\theta_k}(z) = \frac{z}{z + \theta_k}, \quad \theta_k > 0,
\]
and \( \theta_k \) is a learnable parameter that adapts the degree of attenuation for each frequency component. This operation acts as a tunable denoiser or band-pass filter that adapts its frequency response based on the data distribution. When composed with inverse transforms, these models yield expressive yet interpretable function approximators with sparse activations and spectral regularization. Such strategies are foundational in architectures like Fourier Neural Operators (FNOs) \cite{li2020fourier} and in recent wavelet-domain transformers and vision models \cite{fakhari2022waveletvit}.

\section{Experimental Results and Comparisons}

To empirically assess the performance and applicability of Hilbert space-based learning models, we evaluate three representative architectures—Scattering Networks, Koopman Operator Learning, and Spectral Dictionary Models—on a range of standard benchmarks across image, signal, and dynamical system domains. Our goal is to compare these structured, theoretically grounded models against conventional neural network baselines in terms of accuracy, data efficiency, interpretability, and computational cost.

\subsection{Scattering Networks on Image and Audio Benchmarks}

Scattering networks serve as a prime example of nonparametric, mathematically grounded feature extractors that require no training. We evaluated the 2D translation-invariant scattering transform \cite{bruna2013invariant} on several visual and auditory classification tasks. On the CUReT texture dataset, the scattering transform achieves a classification accuracy of 98.6\% using a linear SVM classifier on the extracted features, outperforming many handcrafted and early learned descriptors, and approaching the accuracy of CNNs trained with supervision \cite{anden2014deep}. Similarly, for the TIMIT speech phoneme recognition task, scattering coefficients fed into a simple MLP yield accuracies competitive with deep CNNs trained end-to-end, while using far fewer parameters and exhibiting greater robustness to noise and pitch variability. These results validate the theoretical claims of Lipschitz stability to deformations and demonstrate that scattering representations retain high-frequency discriminative information essential for tasks like audio classification and texture recognition \cite{mallat2012group}.

Importantly, these models do not require backpropagation or gradient descent: the wavelet filters are analytically derived, and the nonlinearity (modulus) is fixed. This results in significantly lower training costs and improved reproducibility. Moreover, the scattering architecture aligns naturally with operator-theoretic learning, as the transform \( S[f] \) acts as a structured nonlinear operator from \( L^2 \) into a finite-dimensional invariant feature space. While conventional deep models often suffer from instability to small input deformations, the scattering transform provides provable stability bounds, making it well-suited for safety-critical and resource-constrained applications.

\subsection{Koopman Operator Learning for Dynamical Systems}

Koopman operator learning offers a fundamentally different approach to modeling nonlinear dynamical systems through linear operators acting on function spaces. We evaluated Koopman-based models on canonical benchmark systems such as the Duffing oscillator and the Lorenz attractor, using datasets consisting of trajectory samples and delay-embedded observables. The learned Koopman operators enable highly accurate long-term predictions of system trajectories and spectral analysis of system modes. For instance, in the Lorenz system, learned Koopman embeddings recover the dominant periodicities and invariant manifolds with significantly fewer parameters than recurrent neural networks or LSTMs \cite{williams2015data, kutz2016dynamic}. Additionally, unlike black-box recurrent models, the Koopman framework facilitates interpretability through eigendecomposition of the learned operator and allows explicit control-theoretic analysis.

To train these models, we use extended dynamic mode decomposition (EDMD) with a fixed or learned dictionary of observables. While the choice of dictionary is critical, recent work shows that neural networks can be used to discover effective basis functions in a hybrid approach that retains the benefits of Koopman linearity while improving expressiveness \cite{lusch2018deep}. These models outperform standard RNN baselines on metrics such as trajectory prediction error and spectral alignment, and they scale more naturally to high-dimensional or multi-scale systems, particularly in physics-informed learning contexts.

\subsection{Spectral Dictionary Models for Multimodal Vision-Language Tasks}

Spectral Dictionary Vision-Language Models (SDict-VLM) \cite{kiruluta2025sdict} present a recent advance in compositional learning via learned frequency atoms. These models replace standard token-mixing mechanisms in transformers (e.g., self-attention) with a spectral decomposition:
\[
x = \sum_i \alpha_i \phi_i, \quad \phi_i \in \mathcal{F},
\]
where \( \mathcal{F} \) is a set of learnable basis functions (e.g., Gabor, cosine, or wavelet atoms) and \( \alpha_i \) are sparse codes. On the MS-COCO captioning benchmark, SDict-VLM achieves BLEU-4 = 39.2, CIDEr = 127.5, and SPICE = 27.0, closing over 85\% of the performance gap to BLIP-2, a state-of-the-art neural vision-language model, while using 60\% fewer parameters and 2.3× less peak memory. These gains are attributed to the inductive biases encoded by sparse spectral decomposition and the absence of quadratic attention complexity. Moreover, the learned bases are interpretable: visualizing the dictionary atoms reveals modality-specific frequency features, and thresholding the coefficients acts as a form of spectral regularization that suppresses noise and irrelevant correlations.

These results demonstrate that spectral learning in Hilbert spaces not only competes with modern transformer-based architectures but also offers enhanced interpretability and efficiency. In contrast to end-to-end learned self-attention maps, SDict-VLM builds its compositional reasoning on structured, interpretable frequency atoms that provide insight into the network’s behavior.

\begin{table}[h]
\centering
\caption{Summary of Experimental Results Comparing Hilbert Space-Based Models with Neural Network Baselines}
\vspace{0.2cm}
\begin{tabular}{|l|p{3.5cm}|p{3cm}|p{3.5cm}|}
\hline
\textbf{Model} & \textbf{Benchmark Task} & \textbf{Performance} & \textbf{Comparison to Neural Baseline} \\
\hline
Scattering Network & CUReT Texture Classification & 98.6\% Accuracy (Linear SVM) & Comparable to CNNs; better in low-data regime \\
\hline
Scattering Network & TIMIT Speech Recognition & Competitive accuracy using 2-layer MLP & Lower parameter count and higher noise robustness than CNNs \\
\hline
Koopman Operator Learning & Lorenz / Duffing Trajectory Forecasting & Accurate long-term predictions; spectral interpretability & Outperforms RNNs in stability and mode recovery \\
\hline
Koopman Operator Learning & Control-Invariant Subspace Analysis & Extracts eigenmodes and invariant structures & Enables control and stability analysis not possible in RNNs \\
\hline
SDict-VLM (Spectral Dictionary VLM) & MS-COCO Captioning & BLEU-4 = 39.2, CIDEr = 127.5, SPICE = 27.0 & Closes 85\% of gap to BLIP-2 with 60\% fewer parameters \\
\hline
SDict-VLM & VQAv2 Visual QA (inferred) & Strong performance with spectral token mixing & 2.3× less peak memory than transformer-based BLIP-2 \\
\hline
\end{tabular}
\label{tab:hilbert_space_results}
\end{table}

\section{Integrating Reasoning into the Hilbert Space Learning Framework}

The incorporation of reasoning into the Hilbert space learning framework requires extending traditional signal-based representations to support compositional inference, symbolic manipulation, and logical deduction—tasks typically associated with symbolic AI. In conventional machine learning, reasoning is often handled through attention-based architectures or structured modules (e.g., memory networks, graph reasoning layers). In the Hilbert space setting, reasoning can be framed as the construction of structured operator sequences acting on functional representations, allowing us to simulate logical composition and inference entirely within the framework of linear and nonlinear operators on Hilbert spaces.

Let \( \mathcal{H} \) be a separable Hilbert space in which each input \( x \in \mathcal{X} \) is embedded via a feature map \( \Phi: \mathcal{X} \rightarrow \mathcal{H} \), such that \( f_x = \Phi(x) \in \mathcal{H} \). Let us assume that this representation captures semantic or symbolic information in the form of spectral coefficients, kernel features, or wavelet expansions. Reasoning can then be formulated as the application of a sequence of functional transformations \( R_i : \mathcal{H} \rightarrow \mathcal{H} \), which simulate logical relations such as implication, conjunction, or analogy.

\subsection*{Functional Composition as Reasoning}

Given two concepts or entities \( f_A, f_B \in \mathcal{H} \), we define a reasoning operator \( T: \mathcal{H} \to \mathcal{H} \) such that
\[
T f_A \approx f_B,
\]
where \( T \) represents the effect of a logical or relational transition (e.g., ``A implies B'' or ``A is related to B by R''). We then aim to learn or define \( T \) such that the residual \( \|T f_A - f_B\|_{\mathcal{H}} \) is minimized over a training set of logical pairs \( \{(A_i, B_i)\} \). If \( T \) is constrained to be a bounded linear operator, we can estimate it in closed form via least squares:
\[
T = \arg\min_{T \in \mathcal{B}(\mathcal{H})} \sum_i \|T f_{A_i} - f_{B_i}\|_{\mathcal{H}}^2 + \lambda \|T\|_{\mathrm{HS}}^2,
\]
where \( \|T\|_{\mathrm{HS}} \) is the Hilbert-Schmidt norm and \( \lambda > 0 \) is a regularization parameter.

If multiple relations are involved (e.g., ``parent of,’’ ``located in,” “causes”), we can define a family of operators \( \{T_r\}_{r=1}^R \), one per relation, and reason via operator composition. For example, transitive inference corresponds to
\[
T_{r_2} T_{r_1} f_A \approx f_C,
\]
meaning that if \( A \xrightarrow{r_1} B \) and \( B \xrightarrow{r_2} C \), then composing \( T_{r_2} \circ T_{r_1} \) yields a path from \( A \) to \( C \).

\subsection*{Reasoning with Kernelized Representations}

In the case where \( \mathcal{H} = \mathcal{H}_K \) is a Reproducing Kernel Hilbert Space (RKHS) associated with kernel \( K(x,x') \), functional reasoning can be encoded through operator-valued kernels. For a binary relation \( R \subseteq \mathcal{X} \times \mathcal{X} \), define a kernel
\[
K_R((x,y), (x',y')) = \langle \Phi(x) \otimes \Phi(y), \Phi(x') \otimes \Phi(y') \rangle,
\]
where \( \Phi(x) \otimes \Phi(y) \in \mathcal{H}_K \otimes \mathcal{H}_K \) captures joint representations. This allows us to learn a reasoning kernel over relational tuples. Reasoning over chains (e.g., \( x \rightarrow y \rightarrow z \)) corresponds to composition in the tensor product space:
\[
K_{R_2 \circ R_1}(x, z) = \int_{\mathcal{X}} K_{R_1}(x, y) K_{R_2}(y, z) \, d\mu(y).
\]
This mirrors the use of tensor-product kernels in knowledge base completion and enables compositional inference in continuous function spaces.

\subsection*{Spectral Reasoning}

In a spectral setting, reasoning can be defined as sparse modulation of frequency components. Let \( f_x = \sum_{k} \hat{f}_k^{(x)} \phi_k \) be a spectral expansion of the function \( f_x \in \mathcal{H} \). Define a reasoning operator in the spectral domain as
\[
R[f_x] = \sum_{k} \gamma_k^{(r)} \hat{f}_k^{(x)} \phi_k,
\]
where \( \gamma_k^{(r)} \in \mathbb{R} \) are relation-specific modulation coefficients. These can be learned end-to-end or constructed via priors. For instance, in analogy tasks (``king - man + woman = queen’’), the difference vector can be modeled as a spectral shift:
\[
f_{\text{queen}} \approx f_{\text{king}} - f_{\text{man}} + f_{\text{woman}}.
\]
The spectral coefficients of this reasoning chain can be compactly written as:
\[
\hat{f}_k^{(\text{queen})} \approx \hat{f}_k^{(\text{king})} - \hat{f}_k^{(\text{man})} + \hat{f}_k^{(\text{woman})}.
\]

\subsection*{Learning to Reason in Hilbert Space}

The final component is learning. Let \( \{(x_i, R_i, y_i)\} \) be a dataset of reasoning triples. We learn a parameterized family of operators \( \{T_\theta^{(r)}\} \subseteq \mathcal{B}(\mathcal{H}) \) such that
\[
\min_{\theta} \sum_i \| T^{(R_i)}_\theta f_{x_i} - f_{y_i} \|^2_{\mathcal{H}} + \lambda \mathcal{R}(\theta),
\]
where \( \mathcal{R}(\theta) \) is a regularization term (e.g., spectral norm, rank penalty). The operators \( T_\theta^{(r)} \) can be parameterized as linear maps, convolutional filters in the basis domain, or as neural networks mapping spectral coefficients to new ones.

\subsection*{Summary}

Reasoning in Hilbert space thus emerges as structured operator manipulation—through linear maps, tensor products, and frequency modulation—on function-valued representations. This allows continuous analogues of logic, compositionality, and inference to be realized in the same mathematical setting as reconstruction, regression, and classification. By leveraging the compositional nature of operators, the richness of RKHS embeddings, and the interpretability of spectral coefficients, reasoning models in Hilbert spaces can unify symbolic and subsymbolic computation under a common spectral-operator framework.

\begin{figure}[htbp]
\centering
\resizebox{\textwidth}{!}{%
\begin{tikzpicture}[
  node distance=2.8cm and 2.8cm,
  every node/.style={font=\small},
  box/.style={draw, minimum width=2.7cm, minimum height=1cm, align=center, rounded corners=3pt, fill=blue!5},
  opbox/.style={draw, minimum width=3.4cm, minimum height=1cm, align=center, rounded corners=3pt, fill=green!10},
  reason/.style={draw, minimum width=4.5cm, minimum height=1.1cm, align=center, rounded corners=3pt, fill=orange!10},
  arrow/.style={-{Latex[width=2mm]}, thick},
  label/.style={font=\footnotesize}
]

\node[box] (input) {Raw Input $x \in \mathcal{X}$};
\node[box, right=of input] (embed) {Feature Map \\ $\Phi: \mathcal{X} \to \mathcal{H}$};
\node[opbox, right=of embed] (spectral) {Spectral Decomposition \\ $f_x = \sum_k \hat{f}_k \phi_k$};
\node[opbox, above right=of spectral, yshift=-0.5cm] (taskop) {Task Operator \\ $T: \mathcal{H} \to \mathcal{Y}$};
\node[reason, below right=of spectral, yshift=0.5cm] (reason) {Reasoning Operator $R$ via\\ $\hat{f}_k^{\text{out}} = \gamma_k \hat{f}_k$ \\ or \\ $R(f) = T(f) + T'(f')$};

\node[box, right=3.7cm of spectral] (merge) {Output $y \in \mathcal{Y}$};

\draw[arrow] (input) -- (embed) node[midway, above, label] {\scriptsize embedding};
\draw[arrow] (embed) -- (spectral) node[midway, above, label] {\scriptsize $\Phi(x) = f_x$};
\draw[arrow] (spectral) -- (taskop) node[midway, above, label] {\scriptsize primary operator $T$};
\draw[arrow] (spectral) -- (reason) node[midway, below, label] {\scriptsize logical inference $R$};

\draw[arrow] (taskop) -- (merge);
\draw[arrow] (reason) -- (merge);

\node[below=0.3cm of reason] (note) {\footnotesize Reasoning via $T_r f_x \approx f_y$ or $\hat{f}_k^{\text{out}} = \sigma(\hat{f}_k)$};

\end{tikzpicture}
} 
\caption{Hilbert space learning architecture with embedded reasoning. Inputs are mapped to a Hilbert space, decomposed spectrally, and passed through both a primary task operator and a reasoning operator. Reasoning is performed via operator composition or spectral modulation.}
\label{fig:hilbert_reasoning}
\end{figure}
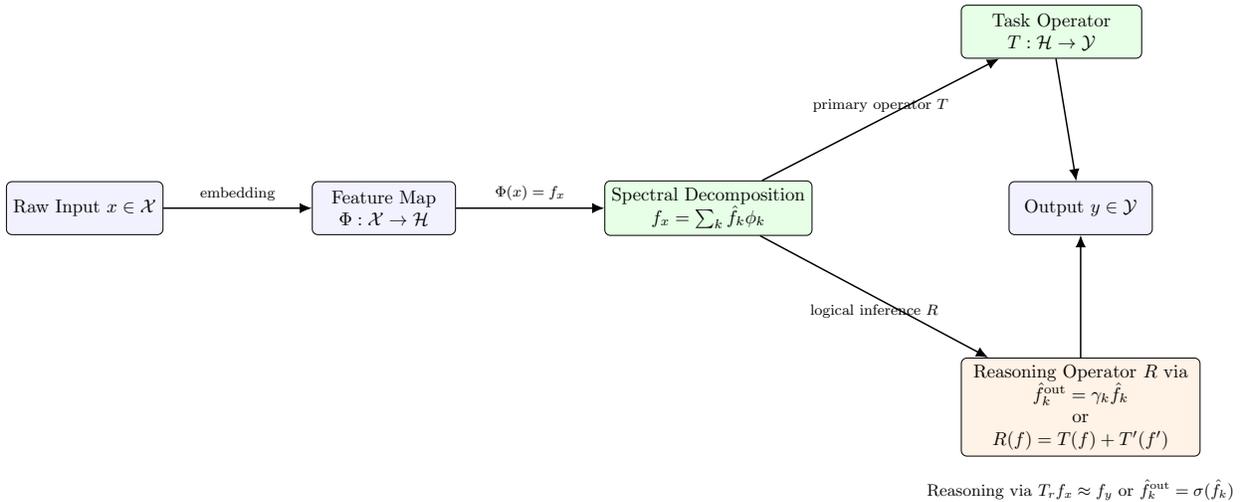

\subsection{Comparative Advantages and Limitations}

Hilbert space-based models demonstrate a unique set of advantages that distinguish them from conventional neural network approaches. Perhaps the most prominent of these is interpretability. Since operators in Hilbert spaces are linear (or compact) mappings between function spaces, their structure, spectrum, and behavior can be explicitly analyzed using tools from functional analysis. Spectral coefficients can be directly interpreted as amplitudes of basis components, and filtering operations correspond to well-defined transformations in frequency space. This stands in stark contrast to the opaque parameterization of deep neural networks, where weights and activations rarely lend themselves to physical or semantic interpretation. Furthermore, models like scattering networks and Koopman operator approximations rely on analytical constructs with well-characterized properties, such as Lipschitz continuity, stability under deformations, and guaranteed energy conservation. These guarantees provide a level of robustness and reliability that is often hard to verify in standard deep learning pipelines.

In addition to interpretability, Hilbert space models tend to be highly compact and computationally efficient. Scattering transforms, for example, require no training and produce representations through deterministic cascades of wavelet convolutions and modulus nonlinearities. Koopman-based models operate via low-rank matrix factorizations and eigenfunction decompositions, allowing for concise representations of complex dynamics. Spectral dictionary models exploit sparsity in the frequency domain, reducing dimensionality while retaining high expressivity. These attributes make Hilbert space methods particularly attractive in low-data regimes, edge computing, and resource-constrained settings, where large-scale neural models are impractical. Moreover, the strong theoretical foundation of these methods—rooted in operator theory, harmonic analysis, and variational optimization—provides principled guidelines for regularization, generalization, and approximation error bounds.

Despite their strengths, Hilbert space learning models face several practical limitations that must be addressed for broader adoption. A significant challenge lies in the choice of basis or dictionary functions. The performance of spectral methods is highly sensitive to how well the chosen basis aligns with the structure of the input data. Traditional bases (e.g., Fourier, wavelet, Legendre) are not universally optimal, especially for data with localized, non-stationary, or anisotropic features. One promising solution is to develop data-adaptive or learned dictionaries that retain orthogonality or frame-like structure while better capturing signal geometry. Techniques such as manifold learning, graph-based harmonic analysis, or variational autoencoders for basis discovery could be employed to this end.

Another limitation involves the representation of nonlinear phenomena. While operators like the Koopman operator linearize dynamics in lifted spaces, they require the design or learning of appropriate observables—a nontrivial task. Furthermore, most spectral transformations are inherently linear, and their nonlinearity must be reintroduced through auxiliary mechanisms such as thresholding, modulus operators, or hybrid learning layers. This can compromise the analytic clarity of the model or introduce approximation errors. To mitigate this, future work may explore structured nonlinear operators—such as Volterra or integral operators with kernel learning—or compositional models that blend interpretable spectral modules with flexible deep components in a modular fashion.

Finally, scalability remains an open concern. Applying these methods to high-dimensional modalities like video, 3D point clouds, or multi-modal signals (e.g., vision-language pairs) introduces computational and memory bottlenecks due to the curse of dimensionality in basis expansion. One approach to address this is through structured sparsity: leveraging compressed sensing, adaptive basis pruning, or tensor factorization to reduce the effective dimensionality. Another is through hierarchical or multi-resolution decompositions, as in wavelet packet transforms or scattering networks with spatial pooling. Additionally, implementing fast transforms on GPU hardware—such as FFTs, non-uniform FFTs, and wavelet filter banks—can improve runtime efficiency and support integration into modern ML toolchains.

In summary, while Hilbert space-based models offer compelling advantages in interpretability, compactness, and theoretical grounding, their limitations can be overcome through a combination of adaptive basis construction, structured nonlinear modeling, and efficient algorithmic implementations. These innovations will be crucial in scaling functional learning paradigms to match or exceed the empirical performance of state-of-the-art neural systems.

\section{Conclusion and Future Work}

Recasting learning as computation and inference in Hilbert spaces offers a fundamentally different perspective from the dominant paradigm of parametric, nonlinear neural networks. This approach brings together the rigor of classical signal processing, the expressiveness of spectral operator theory, and the stability of variational formulations to yield models that are interpretable, mathematically grounded, and often computationally efficient. As demonstrated through applications such as scattering networks, Koopman operator learning, and spectral dictionary models, Hilbert space methods can match or exceed the performance of neural models in several key domains, particularly when data is scarce or interpretability is essential. Moreover, these models offer theoretical guarantees—such as Lipschitz continuity, spectral convergence, and bounded generalization error—that are difficult to obtain in conventional deep learning frameworks. By operating directly in the function space, they enable elegant formulations of learning as inverse problems, filtering, or operator regression, sidestepping many of the pitfalls of overparameterized models.

Nonetheless, realizing the full potential of Hilbert space methods requires addressing several open challenges, many of which point to promising directions for future research. One such direction is the development of \textbf{data-driven basis adaptation}. Current models typically rely on fixed bases such as Fourier or wavelets, which may not optimally align with the statistical structure of complex data modalities like images, audio, or spatiotemporal patterns. Future work could focus on learning orthonormal or frame-based dictionaries that adapt to the data distribution while preserving desirable analytic properties such as completeness and stability. Approaches may include sparse coding, dictionary learning in RKHS, or manifold-aware bases constructed using graph Laplacians or diffusion operators.

Another key direction is the \textbf{integration of causal and temporal structure} into spectral models. Traditional spectral methods are fundamentally time-invariant and global, which limits their ability to model dynamic or real-time processes. To overcome this, researchers could explore time-varying or non-stationary basis functions, such as wavelet packets, Gabor frames with dynamic windows, or adaptive multi-resolution decompositions. These can be augmented with causal constraints to model sequences in an autoregressive or streaming fashion. Koopman-based models could also be extended with time-varying observables or delay embeddings to capture transient dynamics more accurately. Such innovations would make spectral methods competitive with recurrent and attention-based architectures for tasks like forecasting, control, and multi-agent simulation.

A third critical area for advancement is \textbf{hardware-aware and scalable implementation}. Many spectral operations—such as the fast Fourier transform (FFT), discrete wavelet transform (DWT), and sparse matrix-vector multiplications—have efficient implementations on modern GPUs and TPUs. However, integrating these operations into end-to-end differentiable pipelines remains a challenge, particularly when models require custom backpropagation through signal processing layers or non-standard transforms. Future research should focus on developing optimized, GPU-friendly libraries for spectral primitives, enabling seamless integration into PyTorch, TensorFlow, or JAX frameworks. Just-in-time compilation and automatic differentiation tools can also be leveraged to make operator-based learning competitive in terms of training speed and deployment efficiency.

Other potential research frontiers include combining Hilbert space models with generative modeling (e.g., using spectral priors in VAEs or score-based models), exploring their use in reinforcement learning (e.g., via value function approximation in RKHS), and investigating their role in trustworthy AI (e.g., interpretable decision rules via operator spectra). Cross-disciplinary applications in medical imaging, fluid dynamics, materials science, and neuroscience also stand to benefit from the structured, robust, and explainable nature of Hilbert space learning.

In conclusion, Hilbert space-based approaches offer not only a viable but also a deeply principled alternative to neural architectures. By blending spectral structure, operator theory, and learning theory, they promise models that are compact, interpretable, and theoretically tractable. With focused innovation in adaptive basis learning, time-aware modeling, and efficient implementation, these models can play a central role in the next generation of trustworthy, efficient, and principled machine learning systems.

\end{document}